\pdfoutput=1

\documentclass[11pt]{article}

\usepackage[final]{acl}

\usepackage{times}
\usepackage{latexsym}

\usepackage[T1]{fontenc}
\usepackage[utf8]{inputenc}
\usepackage{microtype}
\usepackage{multirow}
\usepackage{booktabs}
\usepackage{caption}
\usepackage{pgfplots}
\pgfplotsset{compat=1.18}
\usepackage{svg}
\usepackage{graphicx}
\usepackage{subcaption}
\usepackage{enumitem}
\usepackage{hyperref}
\hypersetup{
 urlcolor=black
}
\usepackage{inconsolata}

\usepackage{graphicx}

\usepackage{algorithm2e}
\usepackage{amssymb}
\usepackage{amsmath}
\usepackage{colortbl}
\usepackage{xcolor}
\definecolor{cellgray}{gray}{0.85}
\definecolor{textgray}{gray}{0.85}
\definecolor{R_red}{HTML}{F8CECC}
\usepackage{tabularx}
\usepackage[most]{tcolorbox}
\usepackage{array}
\newcolumntype{L}[1]{>{\raggedright\arraybackslash}p{#1}}
\usepackage{makecell}

\title{R$^2$AG: Incorporating Retrieval Information into Retrieval Augmented Generation}

\author{
 \textbf{Fuda Ye\textsuperscript{1}},
 \textbf{Shuangyin Li\textsuperscript{1,\thanks{Corresponding author. The source code is available at \url{https://github.com/yefd/RRAG.git}.}}},
 \textbf{Yongqi Zhang\textsuperscript{2}},
 \textbf{Lei Chen\textsuperscript{2,3}}\\
\\
 \textsuperscript{1}School of Computer Science, South China Normal University\\
 \textsuperscript{2}The Hong Kong University of Science and Technology (Guangzhou)\\
 \textsuperscript{3}The Hong Kong University of Science and Technology\\
 \small{
   \href{mailto:fudayip@m.scnu.edu.cn}{fudayip@m.scnu.edu.cn}, \href{mailto:shuangyinli@scnu.edu.cn}{shuangyinli@scnu.edu.cn}, \href{mailto:yongqizhang@hkust-gz.edu.cn}{yongqizhang@hkust-gz.edu.cn}, \href{mailto:leichen@cse.ust.hk}{leichen@cse.ust.hk}
 }
}

\begin{document}
\maketitle
\begin{abstract}
Retrieval augmented generation (RAG) has been applied in many scenarios to augment large language models (LLMs) with external documents provided by retrievers. However, a semantic gap exists between LLMs and retrievers due to differences in their training objectives and architectures. This misalignment forces LLMs to passively accept the documents provided by the retrievers, leading to incomprehension in the generation process, where the LLMs are burdened with the task of distinguishing these documents using their inherent knowledge. This paper proposes R$^2$AG, a novel enhanced RAG framework to fill this gap by incorporating \textbf{R}etrieval information into \textbf{R}etrieval \textbf{A}ugmented \textbf{G}eneration. Specifically, R$^2$AG utilizes the nuanced features from the retrievers and employs a R$^2$-Former to capture retrieval information. Then, a retrieval-aware prompting strategy is designed to integrate retrieval information into LLMs' generation. Notably, R$^2$AG suits low-source scenarios where LLMs and retrievers are frozen. Extensive experiments across five datasets validate the effectiveness, robustness, and efficiency of R$^2$AG. Our analysis reveals that retrieval information serves as an anchor to aid LLMs in the generation process, thereby filling the semantic gap.\looseness-1

\end{abstract}

\section{Introduction}
\label{sec:intro}

\begin{figure}
    \centering
    \includegraphics[width=1\linewidth]{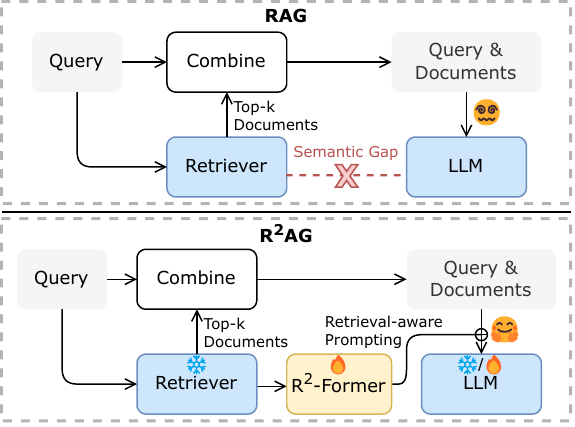}
    \caption{A comparison between RAG and R$^2$AG. R$^2$AG employs a trainable R$^2$-Former to bridge the semantic gap between retrievers and LLMs. Optionally, LLMs can be fine-tuned to understand the retrieval information further.}
    \label{fig:intro}
\end{figure}

Retrieval augmented generation (RAG)~\cite{Lewis2020RetrievalAugmentedGF} significantly enhances the capabilities of large language models (LLMs) by integrating external, non-parametric knowledge provided by retrievers. In RAG framework, the retriever locates and looks up useful documents based on a given query, and then the LLM interacts with these retrieved results to generate a response. The coordination of retrieval and generation achieves impressive performance without additional training. Especially in domain-specific and knowledge-intensive tasks, RAG offers real-time knowledge with high interpretability to LLMs, effectively mitigating the hallucination problem~\cite{Mallen2022WhenNT}.\looseness-1

However, there exists a semantic gap between retrievers and LLMs due to their vastly different training objectives and architectures~\cite{BehnamGhader2022CanRL}. Specifically, retrievers, typically encoder architecture, are designed to retrieve the most relevant documents for a query~\cite{Zhu2023LargeLM}. Conversely, LLMs, generally decoder architecture, are expected to answer questions based on their inherent knowledge or given documents. However, the interaction between retrievers and LLMs in RAG primarily relies on simple text concatenation~\cite{BehnamGhader2022CanRL}. This poor communication strategy will lead to several challenges for LLMs.
\textbf{Externally}, 
it is hard for LLMs to utilize more information from retrievers in separate processes. In RAG, the retrieved documents that only preserve sequential relationships
are unidirectionally delivered to LLMs, and LLMs do not fully understand why retrievers provide the documents. 
Particularly, low-quality documents inevitably appear in retrieved results~\cite{Barnett2024SevenFP}, but LLMs have to accept this noise passively.
\textbf{Internally}, 
it is hard for LLMs to handle all of the retrieved documents with their inherent knowledge. LLMs must process all the results and assess which documents are important, impacting their ability to generate accurate answers~\cite{Wu2024HowED}. Moreover, LLMs face the lost-in-middle problem in overly long documents~\cite{Liu2023LostIT}, leading to further misunderstanding. 

Unfortunately, existing enhanced RAG methods, including pre-processing approaches~\cite{Izacard2022AtlasFL, Yan2024CorrectiveRA, Asai2023SelfRAGLT, Ke2024BridgingTP} and compression-based approaches~\cite{Yan2024CorrectiveRA, Xu2023RECOMPIR, Jiang2023LongLLMLinguaAA}, do not recognize this semantic gap between retrievers and LLMs. They remain to treat retrieval and generation as separate processes and directly add processed or compressed documents into the inputs for LLMs. These strategies ignore the semantic connections necessary for deeper comprehension, which may lead to potentially misleading LLMs even with perfect retrievers.


To address these challenges, it is essential to bridge the semantic gap between retrievers and LLMs. 
As previously mentioned, retrievers can provide high-quality semantic representations that can be beneficial for catching nuanced differences among documents~\cite{Zhao2022DenseTR}. Thus, our intuition is to exploit these semantic representations as additional knowledge, empower LLMs to gain a deeper comprehension of the retrieved documents, and thereby generate more accurate responses.



This paper proposes a cost-effective enhanced RAG framework to incorporate \textbf{R}etrieval information into \textbf{R}etrieval \textbf{A}rgumented \textbf{G}eneration (named R$^2$AG), enhancing LLMs' perception of the key information among retrieved documents. 
Specifically, R$^2$AG adopts an input processing pipeline that transforms semantic representations from a retriever into unified retrieval features.
Then, a trainable R$^2$-Former is employed to capture essential retrieval information. As shown in Figure~\ref{fig:intro}, R$^2$-Former is a pluggable and lightweight model placed between the retriever and the LLM.
Finally, through a retrieval-aware prompting strategy, the LLM receives additional embeddings that contain retrieval information. This strategy aligns the knowledge from retrievers with LLMs without changing the content and order of retrieved documents, thereby relieving information loss.
R$^2$AG offers the flexibility to fine-tune R$^2$-Former alone or both with LLMs. Thus, in R$^2$AG framework, both retrievers and LLMs can be frozen to save computational costs, making R$^2$AG suitable for scenarios with limited resources.
Overall, our contributions are summarized as follows:
\begin{itemize}
    \item We propose R$^2$AG, an enhanced RAG framework, to incorporate retrieval information into retrieval augmented generation. Notably, R$^2$AG is compatible with low-source scenarios where retrievers and LLMs are frozen.
    \item We design a lightweight model, R$^2$-Former, to bridge the semantic gap between retrievers and LLMs. R$^2$-Former can be seamlessly integrated into existing RAG frameworks using open-source LLMs.
    \item We introduce a retrieval-aware prompting strategy to inject retrieval information into the input embeddings, enhancing LLMs' ability to understand relationships among documents without much increase in complexity. 
\end{itemize}
Experimental results demonstrate the superior performance and robustness of R$^2$AG in various scenarios. Our analysis shows that R$^2$AG increases latency by only 0.8\% during inference. Furthermore, it demonstrates that retrieval information anchors LLMs to understand retrieved documents and enhances their generation capabilities.

\section{Related Works}


\subsection{Retrieval Augmented Generation}
Despite being trained on vast corpora, LLMs still struggle with hallucinations and updated knowledge in knowledge-sensitive tasks~\cite{Zhao2023ASO}. RAG~\cite{Lewis2020RetrievalAugmentedGF} is regarded as an efficient solution to these issues by combining a retrieval component with LLMs. In detail, documents gathered by retrievers are bound with the original query and placed into the inputs of LLMs to produce final responses. RAG allows LLMs to access vast, up-to-date data in a flexible way, leading to better performance. Benefiting from the progress of multi-modal alignment techniques~\cite{Li2023BLIP2BL, Zhu2023MiniGPT4EV}, the idea of RAG has been extended to various domains with modality-specific retrievers, including audios~\cite{Koizumi2020AudioCU}, images~\cite{Yasunaga2023RetrievalAugmentedML}, knowledge graphs~\cite{He2024GRetrieverRG}, and so on. Despite its rapid growth, RAG suffers several limitations, such as sensitivity to retrieval results, increased complexity, and a semantic gap between retrievers and LLMs~\cite{Kandpal2022LargeLM, Zhao2024RetrievalAugmentedGF}.\looseness-1


\subsection{Enhanced RAG}
Recent works develop many enhanced approaches based on the standard RAG framework. To directly improve the effectiveness of RAG, REPLUG~\cite{Shi2023REPLUGRB} and Atlas~\cite{Izacard2022AtlasFL} leverage the LLM to provide a supervisory signal for training a better retriever.
However, the noise will inevitably appear in retrieval results~\cite{Barnett2024SevenFP}. 
Recent studies focus on pre-processing the retrieved documents before providing them to LLMs. Techniques such as truncation and selection are effective methods to enhance the quality of ranking lists without modifying the content of documents~\cite{Gao2023RetrievalAugmentedGF, Xu2024ListawareRJ}. CRAG~\cite{Yan2024CorrectiveRA} trains a lightweight retrieval evaluator to exclude irrelevant documents. BGM~\cite{Ke2024BridgingTP} is proposed to meet the preference of LLMs by training a bridge model to re-rank and select the documents. Some studies aim to train small LMs to compress the retrieval documents, thus decreasing complexity or reducing noise. \citet{Jiang2023LongLLMLinguaAA} propose LongLLMLingua to detect and remove unimportant tokens. RECOMP~\cite{Xu2023RECOMPIR} adopts two compressors to select and summarize the retrieved documents. However, the pre-processing methods introduce additional computational costs during inference and may lead to the loss of essential information. 

Notably, the above methods target providing higher-quality retrieval results to LLMs and actually treat retrieval and generation as two distinct processes. This separation fails to bridge the semantic gap between retrievers and LLMs fully. Some approaches~\cite{deng-etal-2023-regavae, sachan2021endtoend} enhance LLM comprehension abilities by incorporating documents into latent representations. However, these methods are typically designed for encoder-decoder LLMs, and constrain their suitability for prevailing decoder-only LLMs. While joint modeling methods~\cite{glass-etal-2022-re2g, 10.5555/3648699.3648950} benefit from the joint optimization of LLMs and retrievers, they need extra training to align semantic spaces, which may hamper the generality of LLMs~\cite{Zhao2024RetrievalAugmentedGF}. Compared with these joint modeling methods, a key difference is that R$^2$AG offers a cost-effective and non-destructive manner to bridge the semantic gap between LLMs and retrievers.




\section{R$^2$AG}

\begin{figure*}
    \centering
    \includegraphics[width=1\linewidth]{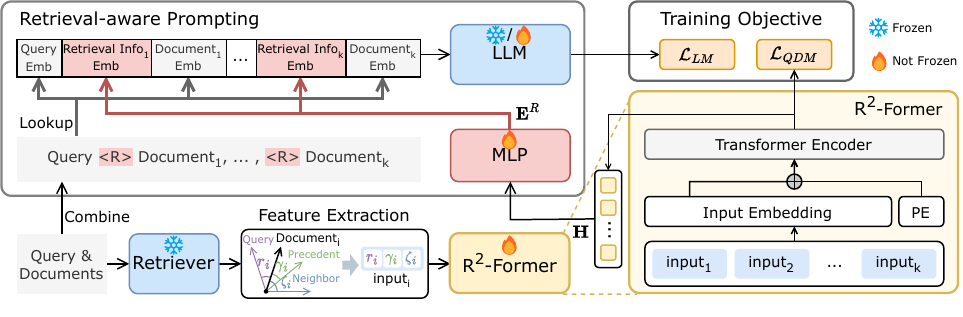}
    \caption{An illustration of R$^2$AG. The R$^2$-Former is designed to extract retrieval features, acting as an information bottleneck between retrievers and LLMs. Through the retrieval-aware prompting strategy, the retrieval information serves as an anchor to guide LLMs during generation. \texttt{``Emb''} is short for embedding, \texttt{``PE''} stands for positional embeddings, and \texttt{``<R>''} denotes the placeholder for retrieval information.}
    \label{fig:overview}
\end{figure*}

\subsection{Problem Formulation and Overview}
RAG involves the task that aims to prompt an LLM to generate answers based on a query and documents returned by a retriever. Formally, given a query $q$ and a list of documents $\mathcal{D}{=}\{d_1, d_2, \cdots , d_k\}$ in preference order ranked by the retriever $f_{\mathbf{R}}$, the LLM, a generator $f_{\mathbf{G}}$, is expected to generate the output $\hat{y}$. The pipeline can be expressed as:
\begin{equation}
    \hat{y}=f_{\mathbf{G}}\left(\text{P}\left(q, \mathcal{D}\right)\right)\text{,}
    \label{eq:rag_generation}
\end{equation}
where $\text{P}$ is a predefined prompt template. It shows the retrievers and LLMs are couple in a simplistic prompt-based method, which will lead to miscommunication and the semantic gap.

Figure~\ref{fig:overview} illustrates the overall framework of R$^2$AG. Initially, given a query and retrieved documents, R$^2$AG processes representations modeled by a retriever into unified-format features. These list-wise features consider nuanced relationships both between the query and documents and among the documents themselves. Then, a R$^2$-Former is designed to capture retrieval information for LLM usage. It allows unified features to interact with each other via self-attention mechanism, enabling it to understand complex dependencies. To integrate retrieval information into the LLM's generation process, R$^2$AG adopts a retrieval-aware prompting strategy to insert the retrieval information into the LLM's input embedding space without causing information loss or increasing much complexity. Besides, R$^2$AG is flexible to be applied in low-source scenarios where LLMs are frozen.


\subsection{Retrieval Feature Extraction}
Before generation, it is necessary to obtain high-quality retrieval features. In R$^2$AG, we first get semantic representations from the retriever $f_{\mathbf{R}}$. Formally, a query $q$ and document $d$ are encoded into representations as $\mathbf{x}^q {=} f_{\mathbf{R}}(q)$ and $\mathbf{x}^d {=} f_{\mathbf{R}}(d)$, respectively. However, these representations can not be directly used because a single representation can not capture interactive features for LLM's generation. Moreover, to suit various retrievers, it is intuitive to transform representations in different spaces into unified format features. 

Inspired by works in retrieval downstream tasks~\cite{Ma2022IncorporatingRI, Ye2024MileCutAM}, we align these representations into retrieval features by computing relevance, precedent similarity, and neighbor similarity scores. Specifically, these scores are calculated by a similarity function such as dot product or cosine similarity. The relevance score $r_i$ is between the query and the $i$-th document and is also used to sort the documents. The precedent and neighbor similarity scores are computed between the $i$-th document representation and its precedent-weighted and adjacent representations, respectively. Detailed formulations are provided in Appendix~\ref{app:retrieval_features}.

Finally, three features are concatenated as input: $\text{input}_i{=}\{r_i, \gamma _i, \zeta _i\}$, representing relevance, precedent similarity, and neighbor similarity. Then, the feature list $\{\text{input}_i\}_{i=1}^{k}$ is then fed into R$^2$-Former to further exploit retrieval information.

\subsection{R$^2$-Former}
Inspired by \citet{Li2023BLIP2BL}, we propose the R$^2$-Former as the trainable module that bridges between retrievers and LLMs. As shown in the right side of Figure~\ref{fig:overview}, R$^2$-Former is a pluggable Transformer-based model that accepts list-wise features as inputs and outputs retrieval information. 

To better comprehend list-wise features from retrievers, we employ an input embedding layer to linearly transform input features into a higher dimension space. Positional embeddings are then added before attention encoding to maintain sequence awareness. Then, a Transformer~\cite{Vaswani2017AttentionIA} encoder is utilized to exploit the input sequences, which uses a self-attention mask where each position's feature can attend to other positions. Formally, for an input list $\{\text{input}_i\}_{i=1}^{k}$, the process is formulated by:
\begin{equation}
    \mathbf{H}=
    f_{att}
    \left[f_{\rightarrow h_1}\left({\left\{\text{input}_i\right\}_{i=1}^k}\right)\text{+}\mathbf{p}\right],
    \label{eq:R2-Former}
\end{equation}
where $f_{att}$ is the Transformer encoder with $h_1$ hidden dimension, $f_{\rightarrow h_1}$ is a linear mapping layer, and $\mathbf{p} \in \mathbb{R}^{k \times h_1}$ represents trainable positional embeddings. The output embeddings $\mathbf{H}\in \mathbb{R}^{k \times {h_1}}$ thus contain the deeper retrieval information and will be delivered to the LLM's generation. 


\subsection{Retrieval-Aware Prompting}
In the generation process, it is crucial for the LLM to utilize the retrieval information effectively. As shown in the upper part of Figure~\ref{fig:overview}, we introduce a retrieval-aware prompting strategy that injects the retrieval information extracted by R$^2$-Former into the LLM's generation process.

First, we employ a projection layer to linearly transform the retrieval information into the same dimension as the token embedding layer of the LLM. Formally, this is represented as:
\begin{equation}
    \mathbf{E}^R=f_{\rightarrow h_2}(\mathbf{H})=\{\mathbf{e}_i^R\}_{i=1}^k,
    \label{eq:prompting_mlp}
\end{equation}
where $f_{\to h_2}$ is a linear projection layer via an MLP layer, and $h_2$ is the dimension of LLM's token embedding layer. 

Then, we tokenize the query and documents using LLM's tokenizer and convert them into embeddings. For example, a document $d$ is tokenized into $\boldsymbol{t}^{d}{=}\{t_j^d\}_{j=1}^{n_d}$, where $t_j^{d}$ is the $j$-th token in the document, $n_{d}$ is the number of tokens in the document $d$. And the token embeddings can be transformed by a lookup in the token embedding layer. The process can be expressed as:
\begin{equation}
    \mathbf{E}^{d}=f_{emb}\left(\boldsymbol{t}^d\right)=\{\mathbf{e}^d_j\}_{j=1}^{n_{d}},
    \label{eq:emb_lookup}
\end{equation}
where $f_{emb}$ is the token embedding layer of the LLM, and $\mathbf{E}^{d}\in \mathbb{R}^{n_d \times h_2}$ is the embeddings of document $d$. A similar process is applied to obtain the query embeddings $\mathbf{E}^q=\{\mathbf{e}^q_j\}_{j=1}^{n_q}$, where $n_q$ is the number of query tokens. 

For nuanced analysis of each document, the corresponding retrieval information embeddings are then prepended to the front of each document's embeddings. They are external knowledge and function as an anchor, guiding the LLM to focus on useful documents. The final input embeddings can be arranged as:
\begin{equation}
\tiny
\mathbf{E}=[
\underbrace{\mathbf{e}_{1}^q, \cdots, \mathbf{e}_{n_q}^q}_{\text{query}}
, \mathbf{e}^{R}_1, 
\underbrace{\mathbf{e}_{1}^{d_1}, \cdots, \mathbf{e}_{n_{d_1}}^{d_1}}_{\text{document}_1}
, \cdots, 
\mathbf{e}^{R}_k, 
\underbrace{\mathbf{e}_{1}^{d_k}, \cdots, \mathbf{e}_{n_{d_k}}^{d_k}}_{\text{document}_k}
],
\label{eq:prompting_concat}
\end{equation}
where $\mathbf{e}_i^R$ denotes the retrieval information embedding for the $i$-th document. In this way, the retrieval information of corresponding document can be well mixed, reducing the burden of the LLM to process all documents. Finally, we can get the responses by:
\begin{equation}
    \hat{y}=f_{\mathbf{G}}(\mathbf{E}),
    \label{eq:generation}
\end{equation}
where $\hat{y}$ represents the LLM-generated results. Notably, this part simplifies the instruction prompt, and detailed descriptions and prompt templates can be found in Appendix~\ref{app:prompt_templates}.

\subsection{Training Strategy}
As the interdependence of retrieval and generation, we integrate R$^2$-Former training and LLM alignment into one stage. The joint training allows R$^2$-Former to better understand list-wise features from the retriever, ensuring retrieval information can be deeply interpreted by the LLM. 

For R$^2$-Former training, we perform a query-document matching (QDM) task that enforces R$^2$-Former to learn the relevance relationships from list-wise features. In detail, it is a binary classification task that asks to model each document's relevance to the query. The formula for prediction is as follows:
\begin{equation}
    \begin{aligned}
     \hat{\mathbf{s}}=f_{\rightarrow 1}(\mathbf{H})=\{\hat{s}_i\}_{i=1}^k,
    \label{eq:rel_pred}
    \end{aligned}
\end{equation}
where $f_{\rightarrow 1}$ is a binary classification head that outputs the relevance predictions $\hat{\mathbf{s}}$. Supporting $\mathbf{s} {=} \{s_i\}^k_{i=1}$ are the ground-truth labels for documents, we use cross-entropy as the loss function, defined as:
\begin{equation}
\small
    \mathcal{L}_{QDM}(\mathbf{s},\hat{\mathbf{s}})={-}\sum_{i=1}^k s_i\log(\hat{s}_i){+}(1{-}s_i)\log(1{-}\hat{s}_i).
    \label{eq:QDM_loss}
\end{equation}

For LLM alignment, we utilize the language modeling (LM) task, which involves learning to generate subsequent tokens based on the preceding context and retrieval information. The language modeling loss $\mathcal{L}_{LM}$ aims to maximize the log-likelihood of the tokens, rewarding the LLM for predicting subsequent words correctly.

The joint training involves instruction fine-tuning with a linear combination of QDM and LM tasks. The final loss is expressed as:
\begin{equation}
    \mathcal{L}=\mathcal{L}_{QDM}{+}\mathcal{L}_{LM}.
    \label{eq:final_loss}
\end{equation}

Notably, R$^2$AG offers the flexibility to train the R$^2$-Former solely while freezing the LLM or to train both together for a deeper understanding of retrieval information. The decision represents a trade-off between lower computational costs and higher accuracy in real-world scenarios.







\section{Experiments}

\subsection{Datasets and Metrics}
We evaluate R$^2$AG on five datasets: Natural Questions (NQ)~\cite{Kwiatkowski2019NaturalQA}, HotpotQA~\cite{Yang2018HotpotQAAD}, MuSiQue~\cite{Trivedi2021MM}, 2WikiMultiHopQA (2Wiki)~\cite{Ho2020ConstructingAM}, and DuReader~\cite{He2017DuReaderAC}. For NQ dataset, we utilize NQ-10, NQ-20, and NQ-30 datasets built by \citet{Liu2023LostIT}, which contain 10, 20, and 30 total documents, respectively. DuReader is a multiple documents QA version built by \citet{Bai2023LongBenchAB}. Detailed introduction and statistics are shown in Appendix~\ref{app:dataset_intro}. 

Following \citet{Mallen2022WhenNT, Liu2023LostIT}, we adopt accuracy (Acc) as the evaluation metric for NQ datasets. Following \citet{Bai2023LongBenchAB}, we adopt accuracy (Acc) and F1 score as evaluation metrics for HotpotQA, MuSiQue, and 2Wiki datasets. For DuReader dataset, we measure performance by F1 score and Rouge~\cite{Lin2004ROUGEAP}.

\begin{table*}[ht]
\centering
\begin{tabular}{@{}l|ccc|cc|cc|cc}
\toprule
\multirow{2}{*}{\textbf{Methods}}
 &\textbf{NQ-10} & \textbf{NQ-20} &\textbf{NQ-30} 
 & \multicolumn{2}{c|}{\textbf{HotpotQA}} 
 & \multicolumn{2}{c|}{\textbf{MuSiQue}} 
 & \multicolumn{2}{c}{\textbf{2Wiki}} 
\\
& \textbf{Acc} & \textbf{Acc} & \textbf{Acc}
& \textbf{Acc} & \textbf{F1} 
& \textbf{Acc} & \textbf{F1} 
& \textbf{Acc} & \textbf{F1} 
\\
\midrule
\multicolumn{10}{c}{\textit{\textbf{Frozen LLMs}}} \\
\midrule
LLaMA2$_{7B}$
& \cellcolor{cellgray}0.3898 & -      & -      
& \cellcolor{cellgray}0.2630 & \cellcolor{cellgray}0.0852 
& \cellcolor{cellgray}0.0546 & \cellcolor{cellgray}0.0241 
& \cellcolor{cellgray}0.1205 & \cellcolor{cellgray}0.0634 \\

{LongChat1.5$_{7B}$}
& 0.6045 & \cellcolor{cellgray}0.5782 & \cellcolor{cellgray}0.5198 
& 0.5424 & 0.3231 
& 0.2808 & 0.1276 
& \underline{0.3882} & \underline{0.2253} \\

LLaMA3$_{8B}$  & 0.5141 & 0.4991 & 0.5311 
& 0.5901 & 0.2056 
& 0.2427 & 0.0891 
& \textbf{0.4723} & {0.1952} \\

LLaMA2$_{13B}$ & \underline{0.7684} & -      & -      
& 0.3788 & 0.1000 
& 0.0909 & 0.0446 
& 0.2405 & 0.0898 \\


ChatGPT     
& \textcolor{gray}{0.6886}    & \textcolor{gray}{0.6761}  & \textcolor{gray}{0.6347}  
& \textcolor{gray}{0.6557}  & \textbf{\textcolor{gray}{0.6518}}  
& \underline{\textcolor{gray}{0.3376}}  & \textbf{\textcolor{gray}{0.3321}}  
& \textcolor{gray}{-}  & \textcolor{gray}{-} \\

GPT4      & \textbf{\textcolor{gray}{0.7759}}    & \textbf{\textcolor{gray}{0.7514}} & \textbf{\textcolor{gray}{0.7514}}  
& \textbf{\textcolor{gray}{0.7673}}  & \underline{\textcolor{gray}{0.6026}}  
& \textbf{\textcolor{gray}{0.4853}}  & \underline{\textcolor{gray}{0.3270}}  
& \textcolor{gray}{-}  & \textcolor{gray}{-} \\

\midrule
CoT             & 0.4482 & 0.6026 & 0.5631 & 0.2365 & 0.1028 & 0.0626 & 0.0412 & 0.1627 & 0.0969 \\
RECOMP          & 0.0169 & 0.2222 & 0.1977 & 0.2388 & 0.0265 & 0.0830 & 0.0156 & 0.2666 & 0.0329 \\
CRAG            & 0.3974 & 0.6441 & 0.6347 & 0.1194 & 0.0360 & 0.0262 & 0.0047 & 0.0768 & 0.0422 \\
\small{LongLLMLingua}   & 0.3635 & -      & -      & 0.4174 & 0.1178 & 0.1939 & 0.0477 & 0.2374 & 0.0888 \\

R$^2$AG& 0.6930 & \underline{0.7062} & \underline{0.6704} & 
\underline{0.6675} & {0.3605} & 
0.1864 & {0.1687} & 
{0.3342} & \textbf{0.3452} \\
\midrule
\multicolumn{10}{c}{\textit{\textbf{Fine-tuned LLMs}}} \\

\midrule
Self-RAG    & 0.1883 & -      & -      & 0.2475 & 0.1236 & 0.0701 & 0.0378 & 0.2611 & 0.1389 \\
RAFT        & 0.7514 & 0.8041 & 0.7307 & 0.7349 & \textbf{0.3172} & \textbf{0.2529} & 0.1502 & \textbf{0.7555} & 0.4869 \\
R$^2$AG+RAFT& \textbf{0.8192} & \textbf{0.8060} & \textbf{0.7458} & \textbf{0.7351} & 0.3056 & 0.2295 & \textbf{0.1533} & 0.7444 & \textbf{0.6351} \\
\bottomrule
\end{tabular}
\caption{Main results on four English datasets. All enhanced RAG methods utilize the same foundation LLMs, with \colorbox{cellgray}{results} marked in gray background indicating the performance of these foundation LLMs. \textcolor{gray}{Results} in gray represent the performance of closed-source LLMs. \textbf{Results} in bold and \underline{results} in underlined mean the best and second-best performance among current classified methods.}
\label{tab:results_en}
\end{table*}

\begin{table}[h]
\centering
\begin{tabular}{l|>{\centering\arraybackslash}p{1.6cm}>{\centering\arraybackslash}p{1.6cm}}
\toprule
\multirow{2}{*}{\textbf{Methods}}
 & \multicolumn{2}{c}{\textbf{DuReader}} 
\\
& \textbf{F1} & \textbf{Rouge} 
\\
\midrule
\multicolumn{3}{c}{\textit{\textbf{Frozen LLMs}}} \\
\midrule
LongChat1.5$_{7B}$& 0.0914 & 0.1181 \\
Qwen1.5$_{0.5B}$& \cellcolor{cellgray}0.1395 & \cellcolor{cellgray}\underline{0.1656}  \\
Qwen1.5$_{1.8B}$& \textbf{0.1533} & 0.1570 \\
InternLM2$_{1.8B}$& 0.1330 & 0.1391 \\
\midrule
R$^2$AG        & \underline{0.1510} & \textbf{0.1663} \\
\midrule
\multicolumn{3}{c}{\textit{\textbf{Fine-tuned LLMs}}} \\
\midrule
RAFT        & 0.2423 & \textbf{0.2740} \\
R$^2$AG+RAFT   & \textbf{0.2507} & 0.2734 \\
\bottomrule
\end{tabular}
\caption{Performance comparison on DuReader dataset.} 
\label{tab:results_zh}
\end{table}

\subsection{Baselines}
To fully evaluate R$^2$AG, we compared two types of methods: standard RAG using various LLMs, and enhanced RAG using the same foundation LLM. 

First, we evaluate standard RAG baselines where LLMs generate responses given the query prepended with retrieved documents. For English datasets, we use several open-source LLMs, including LLaMA2$_{7B}$, LLaMA2$_{13B}$, LLaMA3$_{8B}$~\cite{Touvron2023Llama2O}, and LongChat1.5$_{7B}$~\cite{li_how_2023}. Besides, we adopt ChatGPT~\cite{Ouyang2022TrainingLM} and GPT4~\cite{Achiam2023GPT4TR} as baselines of closed-source LLMs. For the Chinese dataset, we employ Qwen1.5$_{0.5B}$, Qwen1.5$_{1.8B}$~\cite{Bai2023QwenTR} and InternLM2$_{1.8B}$~\cite{Cai2024InternLM2TR}.

Secondly, we experiment with several methods that can enhance RAG, including CoT~\cite{Wei2022ChainOT}, RECOMP~\cite{Xu2023RECOMPIR}, CRAG~\cite{Yan2024CorrectiveRA}, Self-RAG~\cite{Asai2023SelfRAGLT}, LongLLMLingua~\cite{Jiang2023LongLLMLinguaAA}, and RAFT~\cite{Zhang2024RAFTAL}. For NQ-10, HotpotQA, MuSiQue, and 2Wiki datasets, we use LLaMA2$_{7B}$ as the foundation LLM for enhanced RAG methods, which has a maximum context length of 4k tokens. For NQ-20 and NQ-30 datasets, LongChat1.5$_{7B}$ is selected as the foundation LLM, which extends the context window to 32k tokens. For DuReader dataset, Qwen1.5$_{0.5B}$ is the foundation LLM, also with a maximum context length of 32k tokens.

These methods were categorized into two groups -- frozen and fine-tuned -- based on whether they require training the LLMs.

The implementation details are in Appendix~\ref{app:imp_details}. 

\subsection{Main Results}
Table~\ref{tab:results_en} and Table~\ref{tab:results_zh} provide the main results. We can obtain the following conclusions: 

(1) Compared with foundation LLMs using standard RAG, R$^2$AG can significantly increase performance. Even in multi-hot datasets, R$^2$AG improves LLMs' ability for complex reasoning. In DuReader dataset, with a token length of 16k, R$^2$AG remains effective, demonstrating its robustness and efficiency in handling extensive text outputs. These results indicate that R$^2$AG effectively enables LLMs to better understand the retrieval information and boosts their capabilities in handling provided documents. 
(2) Compared with other LLMs using standard RAG, R$^2$AG generally achieves better performance except for closed-source LLMs. GPT4 shows superior results in most datasets, establishing it as a strong baseline. Notably, R$^2$AG excels ChatGPT in NQ and HotpotQA datasets. Using LLaMA2$_{7B}$ as the foundational LLM, R$^2$AG competes well with LLaMA3$_{8B}$ and LLaMA2$_{13B}$ across most metrics.
(3) It is clear that R$^2$AG significantly surpasses other enhanced RAG methods in most results, underscoring the importance of incorporating retrieval information. Although CRAG has a good result in NQ datasets, its performance significantly declines in multi-hop datasets. That is because CRAG's simplistic approach of filtering out documents irrelevant to the query can omit crucial connections needed for understanding complex queries. Additionally, our method outperforms compression-based methods (RECOMP and LongLLMLingua). Our case studies reveal their poor performance is mainly because the coordination between the compressors and LLMs tends to result in substantial information loss and even severe hallucinations.
(4) RAFT can significantly improve the performance. When combined with R$^2$AG, the results are the best overall, suggesting that a deeper understanding acquired through training benefits generation capabilities.



\begin{table}[t]
\small
\centering
\begin{tabular}{l|c|c}
\toprule
 \multirow{2}{*}{\textbf{Methods}} 
 & \multicolumn{1}{c|}{\textbf{NQ-10}}
 & \multicolumn{1}{c}{\textbf{NQ-20}}
 \\
 & LLaMA2$_{7B}$& LongChat1.5$_{7B}$\\
\midrule
\textbf{R$^2$AG}  & \textbf{0.6930}         & \textbf{0.7062}               \\
\midrule
w/o $r$       & 0.6761 ($\downarrow$2.45\%)   & 0.6798 ($\downarrow$3.73\%)   \\
w/o $\gamma$      & 0.6723 ($\downarrow$2.99\%)   & 0.6930 ($\downarrow$1.87\%)   \\
w/o $\zeta$     & 0.6252 ($\downarrow$9.78\%)   & 0.6855 ($\downarrow$2.93\%)   \\
w/o $\mathcal{L}_{QDM}$    & 0.6441 ($\downarrow$7.07\%)   & 0.7043 ($\downarrow$0.27\%)   \\ 
\bottomrule
\end{tabular}
\caption{Ablation studies on NQ-10 and NQ-20 datasets.}
\label{tab:ablation}
\end{table}

\begin{figure}[t]
    \centering
    \includegraphics[width=1\linewidth]{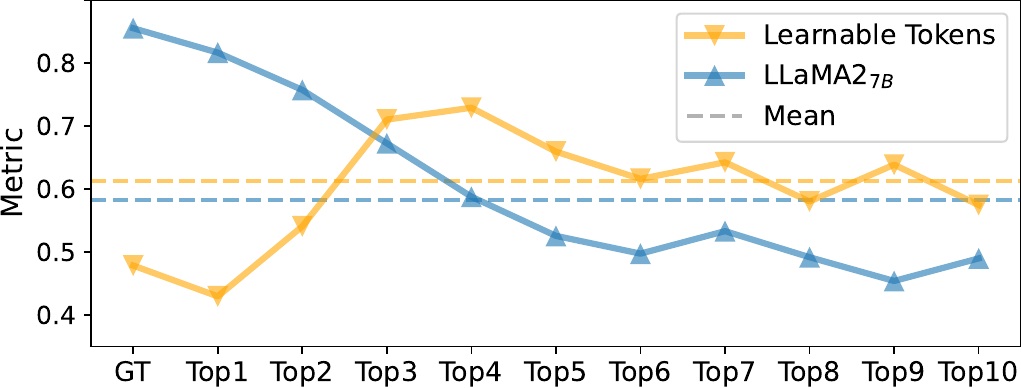}
    \caption{Performance of learnable tokens across different document counts on NQ-10 dataset. \texttt{``GT''} means only retaining ground-true documents.}
    \label{fig:learnable_tokens}
\end{figure}

\begin{figure}[t]
    \centering
    \includegraphics[width=1\linewidth]{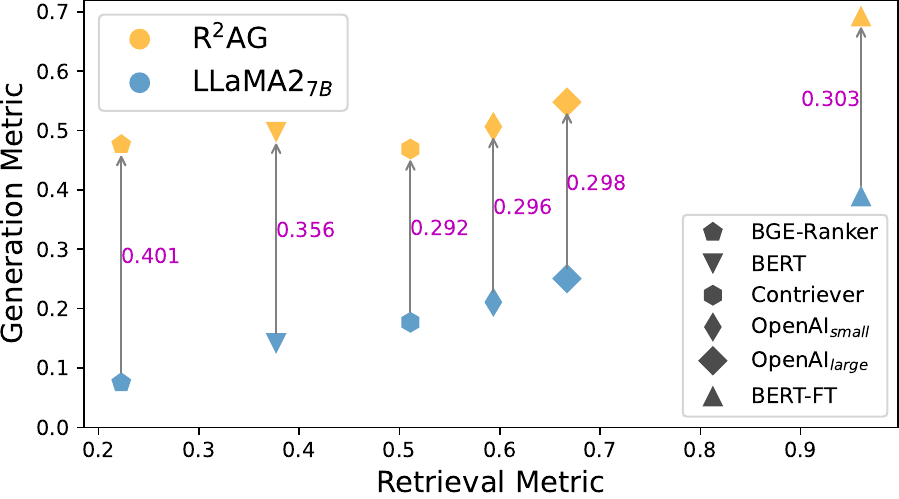}
    \caption{Performance comparison of R$^2$AG with various retrievers on NQ-10 dataset.}
    \label{fig:diff_retriever}
    \vspace{-0.5em}
\end{figure}

\begin{figure}[t]
    \centering
    \includegraphics[width=1\linewidth]{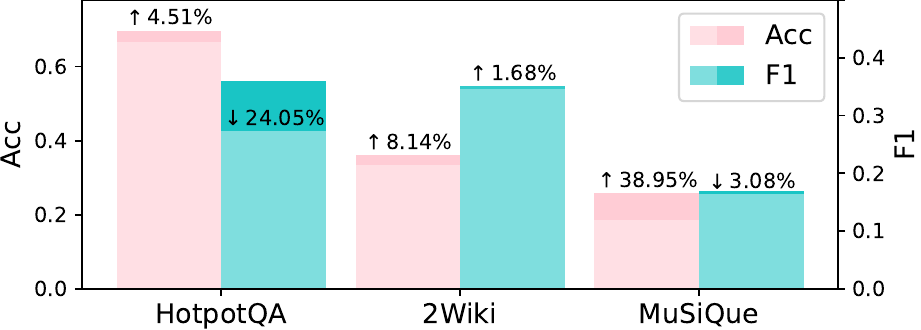}
    \caption{Performance of R$^2$AG$_{7B}$ and R$^2$AG$_{13B}$. Darker parts mean the difference values of R$^2$AG$_{13B}$. }
    \label{fig:diff_llm}
\end{figure}

\subsection{Ablation Studies}

\begin{figure*}[t]
    \centering
    \begin{subfigure}[b]{1\textwidth}
        \centering
        \includegraphics[width=1\linewidth]{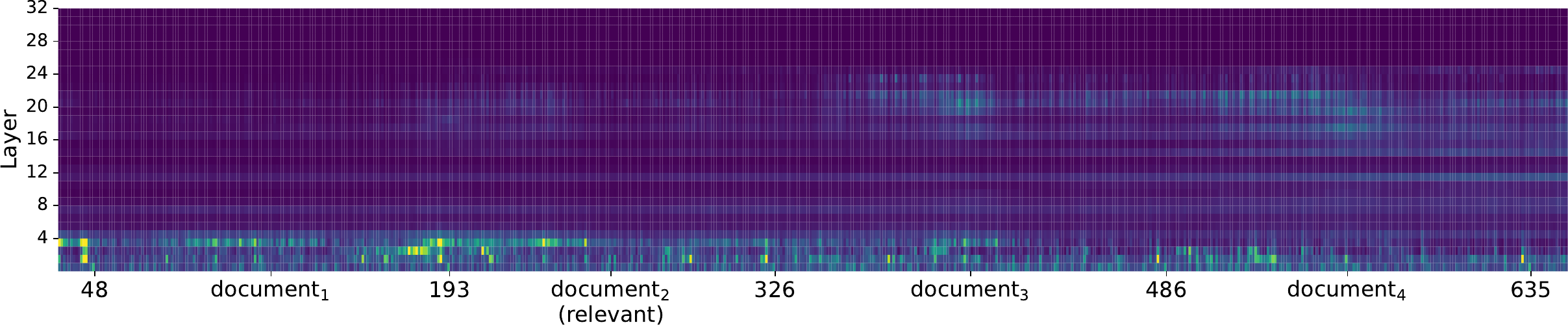}
    \end{subfigure}
    \begin{subfigure}[b]{1\textwidth}
        \includegraphics[width=1\linewidth]{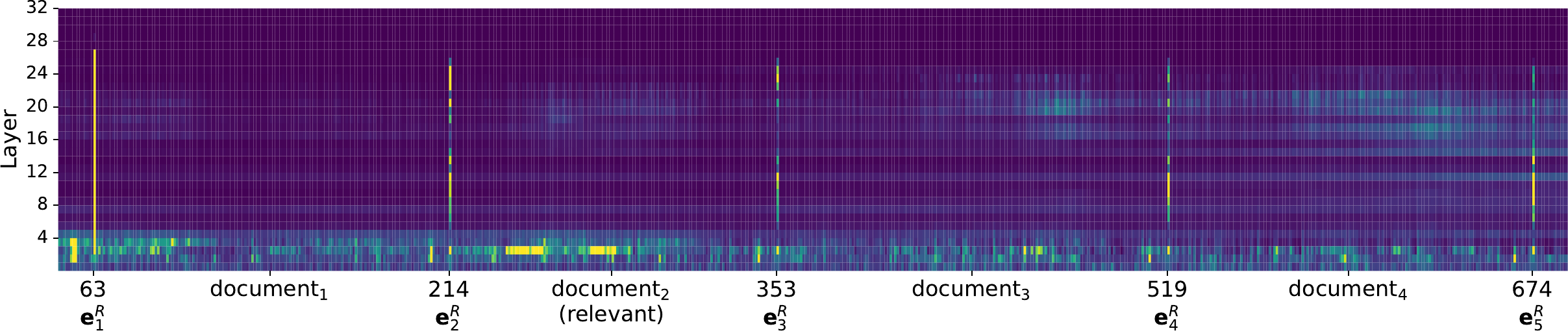}
    \end{subfigure}
    \caption{Heatmaps of self-attention distribution of the last token, broken out by token position (X-axis) and layer (Y-axis). Each attention layer comprises 8 heads, and the attention weights are the mean of all the heads. Darker yellow means higher attention weights. $\mathbf{e}^R_{i}$ is the retrieval information embedding for $i$-th document.} 
    \label{fig:att_distribution}
    \vspace{-0.2em}
\end{figure*}

To demonstrate the effectiveness of R$^2$AG, we create four variants. Specifically, we remove three retrieval features $r,\gamma,\zeta$, individually. For R$^2$-Former, we remove the QDM loss $\mathcal{L}_{QDM}$. We conduct the ablation studies on the NQ-10 and NQ-20 datasets, using LLaMA2$_{7B}$ and LongChat1.5$_{7B}$ as foundation LLMs with results shown in Table~\ref{tab:ablation}.
We can obtain the following observations: First, the performance decreases without any of the three retrieval features, underscoring their effectiveness. The results reveal that utilizing additional retrieval features can help LLMs disentangle irrelevant documents. Secondly, the performance decreases without the QDM loss, showing that the query-document matching task is indeed beneficial for exploiting retrieval information.\looseness-1

To explore the effectiveness of the retrieval-aware prompting strategy, we design an experiment on NQ-10 dataset with various top-$k$ retrieved documents where the retrieval information is set as learnable tokens. This means R$^2$AG only uses these soft prompts without additional features when training and inference. From the results shown in Figure~\ref{fig:learnable_tokens}, we can find that: (1) When retrieval results are largely relevant, with few or no redundant documents, learnable tokens do not aid the LLM and may instead become redundant information for the generation. (2) As the number of documents increases, it is natural to observe a decline performance. Surprisingly, learnable tokens significantly enhance the performance of the LLM. These findings demonstrate that the retrieval-aware prompting strategy effectively assists LLMs in processing multiple documents, especially when those documents include irrelevant information.

\subsection{Discussions}\label{discussion}


\paragraph{The Impact of Performance of Retrievers and LLMs.}
As mentioned in Section~\ref{sec:intro}, the quality of retrieved documents can heavily influence the performance of LLMs in RAG. From the main results, R$^2$AG achieves improvements even when the retrieval performance is poor, as observed in MuSiQue and DuReader datasets. Furthermore, we conduct experiments on NQ-10 dataset with five non-trained retrievers, specifically BGE-Reranker~\cite{Xiao2023CPackPR}, BERT~\cite{Devlin2019BERTPO}, Contriever~\cite{Izacard2022AtlasFL}, and OpenAI Embedding models (small and large)~\cite{Neelakantan2022TextAC}, with 1024, 768, 768, 1536, and 3072 dimensions, respectively. Note that OpenAI Embedding models are closed-source. From the results presented in Figure~\ref{fig:diff_retriever}, we easily observe that a stronger retriever leads to better performance, both standard RAG and R$^2$AG. Importantly, R$^2$AG significantly enhances the effectiveness of LLMs, even when the retrieval performance is poor. 

We conduct experiments on HotpotQA, MuSiQue, and 2Wiki datasets using LLaMA2$_{13B}$ as the foundation LLM. Results shown in Figure~\ref{fig:diff_llm} indicate that R$^2$AG$_{13B}$ outperforms R$^2$AG$_{7B}$, particularly in the accuracy metric. Specially, there is a decline performance in F1 scores for HotpotQA and MuSiQue datasets. We find this primarily because larger LLMs usually tend to output longer answers with explanations (the average response token count in HotpotQA dataset for R$^2$AG$_{7B}$ is 37.44, compared to 49.71 for R$^2$AG$_{13B}$). This tendency also can be observed from the results of ChatGPT and GPT4.

These results reveal that both a stronger LLM and a more effective retriever lead to better performance, validating that R$^2$AG is a genetic method that can be efficiently applied in various scenarios.

\paragraph{The Effect of Retrieval Information.}
For a deeper and more intuitive exploration of how retrieval information improves LLMs' generation, we present a visualization of the self-attention distribution in R$^2$AG compared with standard RAG. In detail, we analyze a case in NQ-10 dataset in which the foundation LLM is LLaMA2$_{7B}$. We extract the self-attention weights in different layers from LLM's outputs and visualize the last token's attention distribution for other tokens. The relevant document is ranked in position 2 in our selected case, while the 1st document is potentially confusing. For a clear illustration, we select attention distribution for tokens in top-4 documents. From Figure~\ref{fig:att_distribution}, it is evident that the retrieval information receives higher attention scores even in deeper layers, and the relevant document can get more attention within 1-4 layers. That means the retrieval information effectively acts as an anchor, guiding the LLM to focus on useful documents. 



\section{Conclusion and Future Work}
This paper proposed a novel enhanced RAG framework named R$^2$AG to bridge the semantic gap between the retrievers and LLMs. By incorporating retrieval information from retrievers into LLMs' generation process, R$^2$AG captures a comprehensive understanding of retrieved documents. Experimental results show that R$^2$AG outperforms other competitors. In addition, the robustness and effectiveness of R$^2$AG are further confirmed by detailed analysis. In future work, more retrieval features could be applied to R$^2$AG framework.

\section*{Limitations}
The following are the limitations associated with R$^2$AG:
First, R$^2$AG depends on the semantic representations modeled by encoder-based retrievers. The suitability of other types of retrievers, such as sparse and cross-encoder retrievers, requires further exploration. Secondly, as mentioned in Section~\ref{discussion}, R$^2$AG relies on the ability of the foundation LLM, and more powerful closed-source LLMs may not be compatible with R$^2$AG. Thirdly, there may be other informative features besides the three retrieval features - relevance, precedent similarity, and neighbor similarity scores. Lastly, R$^2$AG is evaluated on five datasets, of which relevant documents are provided. However, situations where no relevant documents are available need to be considered. R$^2$AG may benefit from integrating techniques like self-RAG to better handle such situations.
\section*{Ethics Statement}
LLMs can generate incorrect and potentially harmful answers. Our proposed method aims to alleviate this issue by providing LLMs with retrieved documents and retrieval information, thereby enhancing LLMs' capability of generation. In the development and execution of our work, we strictly adhered to ethical guidelines established by the broader academic and open-source community. All the datasets and models used in this work are publicly available. No conflicts of interest exist for any of the authors involved in this work.

\section*{Acknowledgments}
This work was supported by Major Program of National Language Committee (WT145-39), Natural Science Foundation of Guangdong (2023A1515012073) and National Natural Science Foundation of China (No. 62006083).\looseness-1


\bibliography{custom}

\begin{thebibliography}{56}
\providecommand{\natexlab}[1]{#1}

\bibitem[{Achiam et~al.(2023)Achiam, Adler, Agarwal, Ahmad, Akkaya, Aleman, Almeida, and et~al}]{Achiam2023GPT4TR}
OpenAI~Josh Achiam, Steven Adler, Sandhini Agarwal, Lama Ahmad, Ilge Akkaya, Florencia~Leoni Aleman, Diogo Almeida, and et~al. 2023.
\newblock \href {https://doi.org/10.48550/arXiv.2303.08774} {Gpt-4 technical report}.

\bibitem[{Asai et~al.(2023)Asai, Wu, Wang, Sil, and Hajishirzi}]{Asai2023SelfRAGLT}
Akari Asai, Zeqiu Wu, Yizhong Wang, Avirup Sil, and Hannaneh Hajishirzi. 2023.
\newblock \href {https://doi.org/10.48550/arXiv.2310.11511} {Self-rag: Learning to retrieve, generate, and critique through self-reflection}.
\newblock \emph{ArXiv}, abs/2310.11511.

\bibitem[{Bai et~al.(2023{\natexlab{a}})Bai, Bai, Chu, Cui, Dang, Deng, Fan, Ge, Han, Huang, Hui, Ji, Li, and et~al}]{Bai2023QwenTR}
Jinze Bai, Shuai Bai, Yunfei Chu, Zeyu Cui, Kai Dang, Xiaodong Deng, Yang Fan, Wenhang Ge, Yu~Han, Fei Huang, Binyuan Hui, Luo Ji, Mei Li, and et~al. 2023{\natexlab{a}}.
\newblock \href {https://doi.org/10.48550/arXiv.2309.16609} {Qwen technical report}.
\newblock \emph{ArXiv}, abs/2309.16609.

\bibitem[{Bai et~al.(2023{\natexlab{b}})Bai, Lv, Zhang, Lyu, Tang, Huang, Du, Liu, Zeng, Hou, and et~al}]{Bai2023LongBenchAB}
Yushi Bai, Xin Lv, Jiajie Zhang, Hong Lyu, Jiankai Tang, Zhidian Huang, Zhengxiao Du, Xiao Liu, Aohan Zeng, Lei Hou, and et~al. 2023{\natexlab{b}}.
\newblock \href {https://doi.org/10.48550/arXiv.2308.14508} {Longbench: A bilingual, multitask benchmark for long context understanding}.
\newblock \emph{ArXiv}, abs/2308.14508.

\bibitem[{Barnett et~al.(2024)Barnett, Kurniawan, Thudumu, Brannelly, and Abdelrazek}]{Barnett2024SevenFP}
Scott Barnett, Stefanus Kurniawan, Srikanth Thudumu, Zach Brannelly, and Mohamed Abdelrazek. 2024.
\newblock \href {https://doi.org/10.48550/arXiv.2401.05856} {Seven failure points when engineering a retrieval augmented generation system}.
\newblock \emph{ArXiv}, abs/2401.05856.

\bibitem[{BehnamGhader et~al.(2022)BehnamGhader, Miret, and Reddy}]{BehnamGhader2022CanRL}
Parishad BehnamGhader, Santiago Miret, and Siva Reddy. 2022.
\newblock \href {https://doi.org/10.48550/arXiv.2212.09146} {Can retriever-augmented language models reason? the blame game between the retriever and the language model}.
\newblock \emph{ArXiv}, abs/2212.09146.

\bibitem[{Cai et~al.(2024)Cai, Cao, Chen, Chen, Chen, Chen, Chen, Chen, Chen, Chu, wen Dong, and et~al}]{Cai2024InternLM2TR}
Zheng Cai, Maosong Cao, Haojiong Chen, Kai Chen, Keyu Chen, Xin Chen, Xun Chen, Zehui Chen, Zhi Chen, Pei Chu, Xiao wen Dong, and et~al. 2024.
\newblock \href {https://doi.org/10.48550/arXiv.2403.17297} {Internlm2 technical report}.
\newblock \emph{ArXiv}, abs/2403.17297.

\bibitem[{Deng et~al.(2023)Deng, Pang, Shen, and Cheng}]{deng-etal-2023-regavae}
Jingcheng Deng, Liang Pang, Huawei Shen, and Xueqi Cheng. 2023.
\newblock \href {https://doi.org/10.18653/v1/2023.findings-emnlp.164} {{R}ega{VAE}: A retrieval-augmented {G}aussian mixture variational auto-encoder for language modeling}.
\newblock In \emph{Findings of the Association for Computational Linguistics: EMNLP 2023}, pages 2500--2510, Singapore. Association for Computational Linguistics.

\bibitem[{Devlin et~al.(2019)Devlin, Chang, Lee, and Toutanova}]{Devlin2019BERTPO}
Jacob Devlin, Ming-Wei Chang, Kenton Lee, and Kristina Toutanova. 2019.
\newblock \href {https://doi.org/10.18653/v1/N19-1423} {{BERT}: Pre-training of deep bidirectional transformers for language understanding}.
\newblock In \emph{Proceedings of the 2019 Conference of the North {A}merican Chapter of the Association for Computational Linguistics: Human Language Technologies, Volume 1 (Long and Short Papers)}, pages 4171--4186, Minneapolis, Minnesota. Association for Computational Linguistics.

\bibitem[{Gao et~al.(2023)Gao, Xiong, Gao, Jia, Pan, Bi, Dai, Sun, Guo, Wang, and Wang}]{Gao2023RetrievalAugmentedGF}
Yunfan Gao, Yun Xiong, Xinyu Gao, Kangxiang Jia, Jinliu Pan, Yuxi Bi, Yi~Dai, Jiawei Sun, Qianyu Guo, Meng Wang, and Haofen Wang. 2023.
\newblock \href {https://doi.org/10.48550/arXiv.2312.10997} {Retrieval-augmented generation for large language models: A survey}.
\newblock \emph{ArXiv}, abs/2312.10997.

\bibitem[{Glass et~al.(2022)Glass, Rossiello, Chowdhury, Naik, Cai, and Gliozzo}]{glass-etal-2022-re2g}
Michael Glass, Gaetano Rossiello, Md~Faisal~Mahbub Chowdhury, Ankita Naik, Pengshan Cai, and Alfio Gliozzo. 2022.
\newblock \href {https://doi.org/10.18653/v1/2022.naacl-main.194} {{R}e2{G}: Retrieve, rerank, generate}.
\newblock In \emph{Proceedings of the 2022 Conference of the North American Chapter of the Association for Computational Linguistics: Human Language Technologies}, pages 2701--2715, Seattle, United States. Association for Computational Linguistics.

\bibitem[{He et~al.(2018)He, Liu, Liu, Lyu, Zhao, Xiao, Liu, Wang, and et~al}]{He2017DuReaderAC}
Wei He, Kai Liu, Jing Liu, Yajuan Lyu, Shiqi Zhao, Xinyan Xiao, Yuan Liu, Yizhong Wang, and et~al. 2018.
\newblock \href {https://doi.org/10.18653/v1/W18-2605} {{D}u{R}eader: a {C}hinese machine reading comprehension dataset from real-world applications}.
\newblock pages 37--46.

\bibitem[{He et~al.(2024)He, Tian, Sun, Chawla, Laurent, LeCun, Bresson, and Hooi}]{He2024GRetrieverRG}
Xiaoxin He, Yijun Tian, Yifei Sun, N.~Chawla, Thomas Laurent, Yann LeCun, Xavier Bresson, and Bryan Hooi. 2024.
\newblock \href {https://doi.org/10.48550/arXiv.2402.07630} {G-retriever: Retrieval-augmented generation for textual graph understanding and question answering}.
\newblock \emph{ArXiv}, abs/2402.07630.

\bibitem[{Ho et~al.(2020)Ho, Duong~Nguyen, Sugawara, and Aizawa}]{Ho2020ConstructingAM}
Xanh Ho, Anh-Khoa Duong~Nguyen, Saku Sugawara, and Akiko Aizawa. 2020.
\newblock \href {https://doi.org/10.18653/v1/2020.coling-main.580} {Constructing a multi-hop {QA} dataset for comprehensive evaluation of reasoning steps}.
\newblock pages 6609--6625.

\bibitem[{Hu et~al.(2021)Hu, Shen, Wallis, Allen-Zhu, Li, Wang, and Chen}]{Hu2021LoRALA}
J.~Edward Hu, Yelong Shen, Phillip Wallis, Zeyuan Allen-Zhu, Yuanzhi Li, Shean Wang, and Weizhu Chen. 2021.
\newblock \href {https://doi.org/10.48550/arXiv.2106.09685} {Lora: Low-rank adaptation of large language models}.
\newblock \emph{ArXiv}, abs/2106.09685.

\bibitem[{Izacard et~al.(2024)Izacard, Lewis, Lomeli, Hosseini, Petroni, Schick, and et~al}]{10.5555/3648699.3648950}
Gautier Izacard, Patrick Lewis, Maria Lomeli, Lucas Hosseini, Fabio Petroni, Timo Schick, and et~al. 2024.
\newblock Atlas: few-shot learning with retrieval augmented language models.
\newblock \emph{J. Mach. Learn. Res.}, 24(1).

\bibitem[{Izacard et~al.(2022)Izacard, Lewis, Lomeli, Hosseini, Petroni, Schick, Yu, Joulin, Riedel, and Grave}]{Izacard2022AtlasFL}
Gautier Izacard, Patrick Lewis, Maria Lomeli, Lucas Hosseini, Fabio Petroni, Timo Schick, Jane~A. Yu, Armand Joulin, Sebastian Riedel, and Edouard Grave. 2022.
\newblock \href {https://doi.org/10.48550/arXiv.2208.03299} {Atlas: Few-shot learning with retrieval augmented language models}.
\newblock \emph{ArXiv}, abs/2208.03299.

\bibitem[{Jiang et~al.(2023)Jiang, Wu, Luo, Li, Lin, Yang, and Qiu}]{Jiang2023LongLLMLinguaAA}
Huiqiang Jiang, Qianhui Wu, Xufang Luo, Dongsheng Li, Chin-Yew Lin, Yuqing Yang, and Lili Qiu. 2023.
\newblock \href {https://doi.org/10.48550/arXiv.2310.06839} {Longllmlingua: Accelerating and enhancing llms in long context scenarios via prompt compression}.
\newblock \emph{ArXiv}, abs/2310.06839.

\bibitem[{Kandpal et~al.(2022)Kandpal, Deng, Roberts, Wallace, and Raffel}]{Kandpal2022LargeLM}
Nikhil Kandpal, H.~Deng, Adam Roberts, Eric Wallace, and Colin Raffel. 2022.
\newblock \href {https://doi.org/10.48550/arXiv.2211.08411} {Large language models struggle to learn long-tail knowledge}.
\newblock In \emph{International Conference on Machine Learning}.

\bibitem[{Ke et~al.(2024)Ke, Kong, Li, Zhang, Mei, and Bendersky}]{Ke2024BridgingTP}
Zixuan Ke, Weize Kong, Cheng Li, Mingyang Zhang, Qiaozhu Mei, and Michael Bendersky. 2024.
\newblock \href {https://doi.org/10.48550/arXiv.2401.06954} {Bridging the preference gap between retrievers and llms}.
\newblock \emph{ArXiv}, abs/2401.06954.

\bibitem[{Kingma and Ba(2014)}]{Kingma2014AdamAM}
Diederik~P. Kingma and Jimmy Ba. 2014.
\newblock \href {https://doi.org/10.48550/arXiv.1412.6980} {Adam: A method for stochastic optimization}.
\newblock \emph{CoRR}, abs/1412.6980.

\bibitem[{Koizumi et~al.(2020)Koizumi, Ohishi, Niizumi, Takeuchi, and Yasuda}]{Koizumi2020AudioCU}
Yuma Koizumi, Yasunori Ohishi, Daisuke Niizumi, Daiki Takeuchi, and Masahiro Yasuda. 2020.
\newblock \href {https://doi.org/10.48550/arXiv.2012.07331} {Audio captioning using pre-trained large-scale language model guided by audio-based similar caption retrieval}.
\newblock \emph{ArXiv}, abs/2012.07331.

\bibitem[{Kwiatkowski et~al.(2019)Kwiatkowski, Palomaki, Redfield, Collins, Parikh, Alberti, Epstein, Polosukhin, and et~al}]{Kwiatkowski2019NaturalQA}
Tom Kwiatkowski, Jennimaria Palomaki, Olivia Redfield, Michael Collins, Ankur~P. Parikh, Chris Alberti, Danielle Epstein, Illia Polosukhin, and et~al. 2019.
\newblock \href {https://doi.org/10.1162/tacl_a_00276} {Natural questions: A benchmark for question answering research}.
\newblock \emph{Transactions of the Association for Computational Linguistics}, 7:453--466.

\bibitem[{Lewis et~al.(2020)Lewis, Perez, Piktus, Petroni, Karpukhin, Goyal, Kuttler, Lewis, and et~al}]{Lewis2020RetrievalAugmentedGF}
Patrick Lewis, Ethan Perez, Aleksandara Piktus, Fabio Petroni, Vladimir Karpukhin, Naman Goyal, Heinrich Kuttler, Mike Lewis, and et~al. 2020.
\newblock \href {https://doi.org/10.48550/arXiv.2005.11401} {Retrieval-augmented generation for knowledge-intensive nlp tasks}.
\newblock \emph{ArXiv}, abs/2005.11401.

\bibitem[{Li et~al.(2023{\natexlab{a}})Li, Shao, Xie, Sheng, Zheng, Gonzalez, Stoica, Ma, and Zhang}]{li_how_2023}
Dacheng Li, Rulin Shao, Anze Xie, Ying Sheng, Lianmin Zheng, Joseph~E. Gonzalez, Ion Stoica, Xuezhe Ma, and Hao Zhang. 2023{\natexlab{a}}.
\newblock \href {https://lmsys.org/blog/2023-06-29-longchat} {How long can open-source {LLMs} truly promise on context length?}

\bibitem[{Li et~al.(2022)Li, Li, Le, Wang, Savarese, and Hoi}]{Li2022LAVISAL}
Dongxu Li, Junnan Li, Hung Le, Guangsen Wang, Silvio Savarese, and Steven C.~H. Hoi. 2022.
\newblock \href {https://doi.org/10.48550/arXiv.2209.09019} {Lavis: A library for language-vision intelligence}.
\newblock \emph{ArXiv}, abs/2209.09019.

\bibitem[{Li et~al.(2023{\natexlab{b}})Li, Li, Savarese, and Hoi}]{Li2023BLIP2BL}
Junnan Li, Dongxu Li, Silvio Savarese, and Steven C.~H. Hoi. 2023{\natexlab{b}}.
\newblock \href {https://doi.org/10.48550/arXiv.2301.12597} {Blip-2: Bootstrapping language-image pre-training with frozen image encoders and large language models}.
\newblock In \emph{International Conference on Machine Learning}.

\bibitem[{Lin(2004)}]{Lin2004ROUGEAP}
Chin-Yew Lin. 2004.
\newblock \href {https://aclanthology.org/W04-1013} {{ROUGE}: A package for automatic evaluation of summaries}.
\newblock In \emph{Text Summarization Branches Out}, pages 74--81, Barcelona, Spain. Association for Computational Linguistics.

\bibitem[{Liu et~al.(2023)Liu, Lin, Hewitt, Paranjape, Bevilacqua, Petroni, and Liang}]{Liu2023LostIT}
Nelson~F. Liu, Kevin Lin, John Hewitt, Ashwin Paranjape, Michele Bevilacqua, Fabio Petroni, and Percy Liang. 2023.
\newblock \href {https://doi.org/10.1162/tacl_a_00638} {Lost in the middle: How language models use long contexts}.
\newblock \emph{Transactions of the Association for Computational Linguistics}, 12:157--173.

\bibitem[{Ma et~al.(2022)Ma, Ai, Wu, Shao, Liu, Zhang, and Ma}]{Ma2022IncorporatingRI}
Yixiao Ma, Qingyao Ai, Yueyue Wu, Yunqiu Shao, Yiqun Liu, M.~Zhang, and Shaoping Ma. 2022.
\newblock \href {https://doi.org/10.1145/3477495.3531998} {Incorporating retrieval information into the truncation of ranking lists for better legal search}.
\newblock \emph{Proceedings of the 45th International ACM SIGIR Conference on Research and Development in Information Retrieval}.

\bibitem[{Mallen et~al.(2023)Mallen, Asai, Zhong, Das, Khashabi, and Hajishirzi}]{Mallen2022WhenNT}
Alex Mallen, Akari Asai, Victor Zhong, Rajarshi Das, Daniel Khashabi, and Hannaneh Hajishirzi. 2023.
\newblock \href {https://doi.org/10.18653/v1/2023.acl-long.546} {When not to trust language models: Investigating effectiveness of parametric and non-parametric memories}.
\newblock In \emph{Proceedings of the 61st Annual Meeting of the Association for Computational Linguistics (Volume 1: Long Papers)}, pages 9802--9822, Toronto, Canada. Association for Computational Linguistics.

\bibitem[{Neelakantan et~al.(2022)Neelakantan, Xu, Puri, Radford, Han, Tworek, Yuan, Tezak, Kim, and et~al}]{Neelakantan2022TextAC}
Arvind Neelakantan, Tao Xu, Raul Puri, Alec Radford, Jesse~Michael Han, Jerry Tworek, Qiming Yuan, Nikolas~A. Tezak, Jong~Wook Kim, and et~al. 2022.
\newblock \href {https://doi.org/10.48550/arXiv.2201.10005} {Text and code embeddings by contrastive pre-training}.
\newblock \emph{ArXiv}, abs/2201.10005.

\bibitem[{Ouyang et~al.(2022)Ouyang, Wu, Jiang, Almeida, Wainwright, Mishkin, Zhang, Agarwal, and et~al}]{Ouyang2022TrainingLM}
Long Ouyang, Jeff Wu, Xu~Jiang, Diogo Almeida, Carroll~L. Wainwright, Pamela Mishkin, Chong Zhang, Sandhini Agarwal, and et~al. 2022.
\newblock \href {https://doi.org/10.48550/arXiv.2203.02155} {Training language models to follow instructions with human feedback}.
\newblock \emph{ArXiv}, abs/2203.02155.

\bibitem[{Paszke et~al.(2019)Paszke, Gross, Massa, Lerer, Bradbury, Chanan, Killeen, Lin, and et~al}]{Paszke2019PyTorchAI}
Adam Paszke, Sam Gross, Francisco Massa, Adam Lerer, James Bradbury, Gregory Chanan, Trevor Killeen, Zeming Lin, and et~al. 2019.
\newblock \href {https://doi.org/10.48550/arXiv.1912.01703} {Pytorch: An imperative style, high-performance deep learning library}.
\newblock \emph{ArXiv}, abs/1912.01703.

\bibitem[{Reimers and Gurevych(2019)}]{Reimers2019SentenceBERTSE}
Nils Reimers and Iryna Gurevych. 2019.
\newblock \href {https://doi.org/10.18653/v1/D19-1410} {Sentence-{BERT}: Sentence embeddings using {S}iamese {BERT}-networks}.
\newblock In \emph{Proceedings of the 2019 Conference on Empirical Methods in Natural Language Processing and the 9th International Joint Conference on Natural Language Processing (EMNLP-IJCNLP)}, pages 3982--3992, Hong Kong, China. Association for Computational Linguistics.

\bibitem[{Sachan et~al.(2021)Sachan, Reddy, Hamilton, Dyer, and Yogatama}]{sachan2021endtoend}
Devendra~Singh Sachan, Siva Reddy, William~L. Hamilton, Chris Dyer, and Dani Yogatama. 2021.
\newblock \href {https://openreview.net/forum?id=5KWmB6JePx} {End-to-end training of multi-document reader and retriever for open-domain question answering}.
\newblock In \emph{Advances in Neural Information Processing Systems}.

\bibitem[{Shi et~al.(2023)Shi, Min, Yasunaga, Seo, James, Lewis, Zettlemoyer, and tau Yih}]{Shi2023REPLUGRB}
Weijia Shi, Sewon Min, Michihiro Yasunaga, Minjoon Seo, Rich James, Mike Lewis, Luke Zettlemoyer, and Wen tau Yih. 2023.
\newblock \href {https://doi.org/10.48550/arXiv.2301.12652} {Replug: Retrieval-augmented black-box language models}.
\newblock \emph{ArXiv}, abs/2301.12652.

\bibitem[{Touvron et~al.(2023)Touvron, Martin, Stone, Albert, Almahairi, Babaei, Bashlykov, Batra, and et~al}]{Touvron2023Llama2O}
Hugo Touvron, Louis Martin, Kevin~R. Stone, Peter Albert, Amjad Almahairi, Yasmine Babaei, Nikolay Bashlykov, Soumya Batra, and et~al. 2023.
\newblock \href {https://doi.org/10.48550/arXiv.2307.09288} {Llama 2: Open foundation and fine-tuned chat models}.
\newblock \emph{ArXiv}, abs/2307.09288.

\bibitem[{Trivedi et~al.(2021)Trivedi, Balasubramanian, Khot, and Sabharwal}]{Trivedi2021MM}
H.~Trivedi, Niranjan Balasubramanian, Tushar Khot, and Ashish Sabharwal. 2021.
\newblock \href {https://doi.org/10.1162/tacl_a_00475} {Musique: Multihop questions via single-hop question composition}.
\newblock \emph{Transactions of the Association for Computational Linguistics}, 10:539--554.

\bibitem[{Vaswani et~al.(2017)Vaswani, Shazeer, Parmar, Uszkoreit, Jones, Gomez, Kaiser, and Polosukhin}]{Vaswani2017AttentionIA}
Ashish Vaswani, Noam~M. Shazeer, Niki Parmar, Jakob Uszkoreit, Llion Jones, Aidan~N. Gomez, Lukasz Kaiser, and Illia Polosukhin. 2017.
\newblock \href {https://doi.org/10.48550/arXiv.1706.03762} {Attention is all you need}.
\newblock In \emph{Neural Information Processing Systems}.

\bibitem[{Wei et~al.(2022)Wei, Wang, Schuurmans, Bosma, hsin Chi, Xia, Le, and Zhou}]{Wei2022ChainOT}
Jason Wei, Xuezhi Wang, Dale Schuurmans, Maarten Bosma, Ed~Huai hsin Chi, F.~Xia, Quoc Le, and Denny Zhou. 2022.
\newblock \href {https://doi.org/10.48550/arXiv.2201.11903} {Chain of thought prompting elicits reasoning in large language models}.
\newblock \emph{ArXiv}, abs/2201.11903.

\bibitem[{Wolf et~al.(2020)Wolf, Debut, Sanh, Chaumond, Delangue, Moi, Cistac, Rault, and et~al}]{Wolf2020Transformers}
Thomas Wolf, Lysandre Debut, Victor Sanh, Julien Chaumond, Clement Delangue, Anthony Moi, Pierric Cistac, Tim Rault, and et~al. 2020.
\newblock \href {https://doi.org/10.18653/v1/2020.emnlp-demos.6} {Transformers: State-of-the-art natural language processing}.
\newblock pages 38--45.

\bibitem[{Wu et~al.(2024)Wu, Xie, Chen, Zhu, Zhang, and Xiao}]{Wu2024HowED}
Siye Wu, Jian Xie, Jiangjie Chen, Tinghui Zhu, Kai Zhang, and Yanghua Xiao. 2024.
\newblock \href {https://doi.org/10.48550/arXiv.2404.03302} {How easily do irrelevant inputs skew the responses of large language models?}
\newblock \emph{ArXiv}, abs/2404.03302.

\bibitem[{Xiao et~al.(2023)Xiao, Liu, Zhang, and Muennighoff}]{Xiao2023CPackPR}
Shitao Xiao, Zheng Liu, Peitian Zhang, and Niklas Muennighoff. 2023.
\newblock \href {https://doi.org/10.48550/arXiv.2309.07597} {C-pack: Packaged resources to advance general chinese embedding}.
\newblock \emph{ArXiv}, abs/2309.07597.

\bibitem[{Xu et~al.(2023)Xu, Shi, and Choi}]{Xu2023RECOMPIR}
Fangyuan Xu, Weijia Shi, and Eunsol Choi. 2023.
\newblock \href {https://doi.org/10.48550/arXiv.2310.04408} {Recomp: Improving retrieval-augmented lms with compression and selective augmentation}.
\newblock \emph{ArXiv}, abs/2310.04408.

\bibitem[{Xu et~al.(2024)Xu, Pang, Xu, Shen, and Cheng}]{Xu2024ListawareRJ}
Shicheng Xu, Liang Pang, Jun Xu, Huawei Shen, and Xueqi Cheng. 2024.
\newblock \href {https://doi.org/10.1145/3589334.3645336} {List-aware reranking-truncation joint model for search and retrieval-augmented generation}.
\newblock In \emph{The Web Conference}.

\bibitem[{Yan et~al.(2024)Yan, Gu, Zhu, and Ling}]{Yan2024CorrectiveRA}
Shi-Qi Yan, Jia-Chen Gu, Yun Zhu, and Zhen-Hua Ling. 2024.
\newblock \href {https://doi.org/10.48550/arXiv.2401.15884} {Corrective retrieval augmented generation}.
\newblock \emph{ArXiv}, abs/2401.15884.

\bibitem[{Yang et~al.(2018)Yang, Qi, Zhang, Bengio, Cohen, Salakhutdinov, and Manning}]{Yang2018HotpotQAAD}
Zhilin Yang, Peng Qi, Saizheng Zhang, Yoshua Bengio, William~W. Cohen, Ruslan Salakhutdinov, and Christopher~D. Manning. 2018.
\newblock \href {https://doi.org/10.18653/v1%2FD18-1259} {Hotpotqa: A dataset for diverse, explainable multi-hop question answering}.
\newblock In \emph{Conference on Empirical Methods in Natural Language Processing}.

\bibitem[{Yasunaga et~al.(2023)Yasunaga, Aghajanyan, Shi, James, Leskovec, Liang, Lewis, Zettlemoyer, and tau Yih}]{Yasunaga2023RetrievalAugmentedML}
Michihiro Yasunaga, Armen Aghajanyan, Weijia Shi, Rich James, Jure Leskovec, Percy Liang, Mike Lewis, Luke Zettlemoyer, and Wen tau Yih. 2023.
\newblock \href {https://doi.org/10.48550/arXiv.2211.12561} {Retrieval-augmented multimodal language modeling}.
\newblock \emph{ArXiv}, abs/2211.12561.

\bibitem[{Ye and Li(2024)}]{Ye2024MileCutAM}
Fuda Ye and Shuangyin Li. 2024.
\newblock \href {https://doi.org/10.1145/3589334.3645349} {Milecut: A multi-view truncation framework for legal case retrieval}.
\newblock In \emph{Proceedings of the ACM Web Conference 2024}, WWW '24, page 1341–1349, New York, NY, USA. Association for Computing Machinery.

\bibitem[{Zhang et~al.(2024)Zhang, Patil, Jain, Shen, Zaharia, Stoica, and Gonzalez}]{Zhang2024RAFTAL}
Tianjun Zhang, Shishir~G. Patil, Naman Jain, Sheng Shen, Matei~A. Zaharia, Ion Stoica, and Joseph~E. Gonzalez. 2024.
\newblock \href {https://doi.org/10.48550/arXiv.2403.10131} {Raft: Adapting language model to domain specific rag}.
\newblock \emph{ArXiv}, abs/2403.10131.

\bibitem[{Zhao et~al.(2024)Zhao, Zhang, Yu, Wang, Geng, Fu, Yang, Zhang, and Cui}]{Zhao2024RetrievalAugmentedGF}
Penghao Zhao, Hailin Zhang, Qinhan Yu, Zhengren Wang, Yunteng Geng, Fangcheng Fu, Ling Yang, Wentao Zhang, and Bin Cui. 2024.
\newblock \href {https://doi.org/10.48550/arXiv.2402.19473} {Retrieval-augmented generation for ai-generated content: A survey}.
\newblock \emph{ArXiv}, abs/2402.19473.

\bibitem[{Zhao et~al.(2022)Zhao, Liu, Ren, and rong Wen}]{Zhao2022DenseTR}
Wayne~Xin Zhao, Jing Liu, Ruiyang Ren, and Ji~rong Wen. 2022.
\newblock \href {https://doi.org/10.1145/3637870} {Dense text retrieval based on pretrained language models: A survey}.
\newblock \emph{ACM Transactions on Information Systems}.

\bibitem[{Zhao et~al.(2023)Zhao, Zhou, Li, Tang, Wang, Hou, Min, Zhang, Zhang, and et~al}]{Zhao2023ASO}
Wayne~Xin Zhao, Kun Zhou, Junyi Li, Tianyi Tang, Xiaolei Wang, Yupeng Hou, Yingqian Min, Beichen Zhang, Junjie Zhang, and et~al. 2023.
\newblock \href {https://doi.org/10.48550/arXiv.2303.18223} {A survey of large language models}.
\newblock \emph{ArXiv}, abs/2303.18223.

\bibitem[{Zhu et~al.(2023{\natexlab{a}})Zhu, Chen, Shen, Li, and Elhoseiny}]{Zhu2023MiniGPT4EV}
Deyao Zhu, Jun Chen, Xiaoqian Shen, Xiang Li, and Mohamed Elhoseiny. 2023{\natexlab{a}}.
\newblock \href {https://doi.org/10.48550/arXiv.2304.10592} {Minigpt-4: Enhancing vision-language understanding with advanced large language models}.
\newblock \emph{ArXiv}, abs/2304.10592.

\bibitem[{Zhu et~al.(2023{\natexlab{b}})Zhu, Yuan, Wang, Liu, Liu, Deng, Dou, and rong Wen}]{Zhu2023LargeLM}
Yutao Zhu, Huaying Yuan, Shuting Wang, Jiongnan Liu, Wenhan Liu, Chenlong Deng, Zhicheng Dou, and Ji~rong Wen. 2023{\natexlab{b}}.
\newblock \href {https://doi.org/10.48550/arXiv.2308.07107} {Large language models for information retrieval: A survey}.
\newblock \emph{ArXiv}, abs/2308.07107.

\end{thebibliography}

\appendix

\begin{table*}[t]
\small
\centering
\begin{tabular}{c|cccccc}
\toprule
\textbf{Datasets} & \textbf{Language} & \textbf{\# Query} & \textbf{\# Train/Test} & \textbf{\# Tokens}& \textbf{\# Rel/Docs}& \textbf{MAP} \\
\midrule
NQ-10 & English & 2655 & 2124/531 & $\sim$2k & 1/10 & 0.9602 \\ 
NQ-20 & English & 2655 & 2124/531 & $\sim$4k & 1/20 & 0.9287 \\ 
NQ-30 & English & 2655 & 2124/531 & $\sim$6k & 1/30 & 0.9215 \\
HotpotQA & English & 97852 & 90447/7405 & $\sim$2k & 2.36/10 & 0.9138 \\
MuSiQue & English & 22355 & 19938/2417 & $\sim$3k & 2.37/20 & 0.5726 \\
2Wiki& English & 180030 & 167454/12576 & $\sim$2k & 2.42/10 & 0.9637 \\
DuReader & Chinese & 200 & 160/40 & $\sim$16k & 1.82/20 & 0.7169 \\
\bottomrule
\end{tabular}
\caption{Statistics of datasets. ``\# Rel/Docs'' denotes the number of relevant documents and the total number of documents for each query. ``MAP'' represents the Mean Average Precision, a common retrieval metric.}
\label{tab:datasets_stat}
\end{table*}

\section{Retrieval Feature Extraction Details}\label{app:retrieval_features}
Formally, the relevance between the query and the $i$-th document is calculated as:
\begin{equation}
 r_i=\text{sim}\left(\mathbf{x}^q, \mathbf{x}^d_i\right),
 \label{eq:rel_sim}
\end{equation}
where $\text{sim}$ is a similarity function such as dot product or cosine similarity, $\mathbf{x}^q$ and $\mathbf{x}^d_i$ are representations of query and $i$-th document, respectively.

The precedent similarity computes the similarity score between case representation and its precedent-weighted representations in the ranking list as follows:
\begin{equation}
    \gamma_i{=}\text{sim}\left(\mathbf{x}^d_i, \sum_{j=1}^{i-1}w_j\cdot \mathbf{x}^d_j\right), w_j{=}\frac{\text{exp}(r_j)}{\sum^{k}_{\ell=1}\text{exp}(r_\ell)},
    \label{eq:pre_sim}
\end{equation}
where $\gamma_i$ is the precedent similarity between $i$-th document and its precedents in the ranking list, and $r_i$ is relevance between the query and $i$-th document.

Neighbor similarity represents the average similarity of $i$-th document to its adjacent documents. Specifically, the neighbor similarity of a case in the ranking list is given by:
\begin{align}
\small
\zeta_i = 
\begin{cases}
\text{sim}(\mathbf{x}^d_1, \mathbf{x}^d_2), & i = 1 \\
[\text{sim}(\mathbf{x}^d_{i{-}1}, \mathbf{x}^d_i)+\text{sim}(\mathbf{x}^d_i, \mathbf{x}^d_{i{+}1})]/2, & i \in [2, k) \\
\text{sim}(\mathbf{x}^d_{k-1}, \mathbf{x}^d_k), & i = k
\end{cases},
\label{eq:nb_sim}
\end{align}
where $\zeta_i$ represents the average similarity of $i$-th document to its adjacent documents. Such that we can get the list-wise features among documents.

\section{Prompt Templates}\label{app:prompt_templates}

In R$^2$AG, retrieval information, we append $k$ special tokens (\texttt{``<R>''}) in front of each document to facilitate the incorporation of retrieval information. These tokens do not carry meaningful semantics but serve as placeholders for the retrieval information within the prompt. This special token facilitates the integration of retrieval information into the generation process.

Table~\ref{tab:prompt_template} shows the prompt templates for R$^2$AG and other baselines. The prompt templates of DuReader dataset can be found in our source code.

\section{Dataset Introduction}\label{app:dataset_intro}
We conduct evaluations on five datasets, including:

\paragraph{Natural Questions (NQ)}~\cite{Kwiatkowski2019NaturalQA} is developed from Google Search and contains questions coupled with human-annotated answers extracted from Wikipedia. Further, \citet{Liu2023LostIT} collect $k{-}1$ distractor documents from Wikipedia that do not contain the answers, where $k$ is the total document number for each question. This dataset has three versions: NQ-10, NQ-20, and NQ-30, with total document numbers of 10, 20, and 30, respectively.

\paragraph{HotpotQA}~\cite{Yang2018HotpotQAAD} is a well-known multi-hop question answering dataset based on Wikipedia. This dataset involves questions requiring finding and reasoning over multiple supporting facts from 10 documents. There are two reasoning types of questions: bridging and comparison. 

\paragraph{MuSiQue}~\cite{Trivedi2021MM} has questions that involve 2-4 hops and six types of reasoning chains. The dataset is constructed through a bottom–up process by carefully selecting and composing single-hop questions. The final answer to each question in the distractor setting is extracted from 20 documents.

\paragraph{2WikiMultiHopQA (2Wiki)}~\cite{Ho2020ConstructingAM} consists of up to 5-hop questions, each associated with 10 documents. Unlike HotpotQA, this dataset needs to evaluate the interpretability of models not only with supporting evidence but also with entity-relation tuples. 

\paragraph{DuReader}~\cite{He2017DuReaderAC} is a Chinese dataset developed based on Baidu Search and Baidu Zhidao. To adapt it for assessing long context ability, for each question, \citet{Bai2023LongBenchAB} arbitrarily select several documents from the total corpus as distractors until each question is associated with 20 candidate documents.

The ground truth labels are provided in original datasets. Detailed statistics can be found in Table~\ref{tab:datasets_stat}.

\begin{table}[h]
\small
\centering
\begin{tabular}{l|L{0.7\linewidth}}
    
\toprule
    
\textbf{Methods}   &  \textbf{Prompts} \\
\midrule
RAG &
Write a high-quality answer for the given question using only the provided search results (some of which might be irrelevant). Only give me the answer and do not output any other words.

[1]\{\#$d_1$\}

[2]\{\#$d_2$\}

...

[$k$]\{\#$d_k$\}

Only give me the answer and do not output any other words.

Question: \{\#$q$\}

Answer:
\\
\midrule

CoT &

Write a high-quality answer for the given question using only the provided search results (some of which might be irrelevant). Only give me the answer and do not output any other words.

[1]\{\#$d_1$\}

[2]\{\#$d_2$\}

...

[$k$]\{\#$d_k$\}

Only give me the answer and do not output any other words.

Question: \{\#$q$\}

Let's think it step by step.
\\
\midrule
\makecell{Comps} & Write a high-quality answer for the given question using only the provided search results (some of which might be irrelevant). Only give me the answer and do not output any other words.

\{\#\textit{Compressed documents}\}

Only give me the answer and do not output any other words.

Question: \{\#$q$\}

Answer:
\\

\midrule

R$^2$AG &

Write a high-quality answer for the given question using only the provided search results (some of which might be irrelevant). Only give me the answer and do not output any other words. The similarity information is provided in front of search results.

[1]similarity: \texttt{<R>}\{\#$d_1$\}

[2]similarity: \texttt{<R>}\{\#$d_2$\}

...

[$k$]similarity: \texttt{<R>}\{\#$d_k$\}

Only give me the answer and do not output any other words. 

Question: \{\#$q$\}

Answer:
\\

\bottomrule
\end{tabular}
\caption{Prompt templates of different methods. ``Comps'' means compression-based methods, including RECOMP and LongLLMLingua. \texttt{``<R>''} is the placeholder for retrieval information.}
\label{tab:prompt_template}
\end{table}

\section{Implementation Details}\label{app:imp_details}
Unlike some works~\cite{Li2023BLIP2BL, Zhu2023MiniGPT4EV} built on LAVIS~\cite{Li2022LAVISAL}, we completely implement R$^2$AG on PyTorch~\cite{Paszke2019PyTorchAI} and Transformers~\cite{Wolf2020Transformers} libraries for easy usage. 

For the retrieval task, we utilize the Sentence-Transformer~\cite{Reimers2019SentenceBERTSE} to fine-tune a BERT~\cite{Devlin2019BERTPO} model as the retriever, which is a siamese dual encoder with shared parameters. The models \texttt{``bert-base-uncased''} and \texttt{``bert-base-chinese''} are used for English datasets and the Chinese dataset, respectively. All retrievers adopt default hyper-parameter settings with 768 embedding dimensions. Cosine similarity is employed as the scoring function for retrieval and feature extraction. The retrieval performance across datasets is shown in Table~\ref{tab:datasets_stat}. Contrary to some works~\cite{Liu2023LostIT, Jiang2023LongLLMLinguaAA} that artificially place ground truth documents in fixed positions, this paper considers that candidate documents are ranked by the retriever to simulate real-world scenarios.

For R$^2$-Former, we determine the learning rate as 2e-4 and dropout as 0.1. The number of attention heads and hidden size in Transformer encoder are 4 and 256, respectively. Adam~\cite{Kingma2014AdamAM} is adopted as the optimization algorithm. 

For LLMs, all methods use default settings and adopt greedy decoding for fair comparison. The ChatGPT version is \texttt{``gpt-3.5-turbo-0125''} with a 16k context window size, and the GPT4 version is \texttt{``gpt-4-turbo-2024-04-09''} with a 128k context window size. In CRAG, the retrieval evaluator only triggered \texttt{\{Correct, Ambiguous\}} actions to next knowledge refinement process as there are at least one relevant document in retrieval results. In the RAFT method, we employ LoRA~\cite{Hu2021LoRALA} to effectively fine-tune LLMs, with LoRA rank set at 16, alpha at 32, and dropout at 0.1.

\end{document}